\documentclass{article} % For LaTeX2e
\usepackage{iclr2027_conference,times}

\usepackage{url}

\usepackage{amsmath,amssymb}
\usepackage{amsmath,amsfonts,bm}

\def\eqref#1{equation~\ref{#1}}
\def\1{\bm{1}}

\DeclareMathAlphabet{\mathsfit}{\encodingdefault}{\sfdefault}{m}{sl}
\SetMathAlphabet{\mathsfit}{bold}{\encodingdefault}{\sfdefault}{bx}{n}

\usepackage{graphicx}
\usepackage{gradient-text}
\usepackage[colorlinks=true,linkcolor=cvprblue,citecolor=cvprblue,urlcolor=cvprblue]{hyperref}
\graphicspath{{Figure/}{./}}
\DeclareGraphicsExtensions{.pdf,.png,.jpg,.jpeg}

\definecolor{cvprblue}{rgb}{0.21,0.49,0.74}

\usepackage{booktabs}
\usepackage{wrapfig}
\usepackage{multirow}
\usepackage{makecell}
\usepackage{array}
\usepackage{pifont}

\providecommand{\cmark}{\ding{51}}
\providecommand{\xmark}{\ding{55}}

\usepackage{xspace}
\newcommand{\ourmethod}{\textsc{RoBoSTAR}\xspace}

\title{\ourmethod: Next-Scale Autoregressive Sign Language Translation for Humanoid Robots}

\author{
Yujia Zeng$^{*}$, Chensheng Peng$^{*}$, Yuxin Chen$^{*}$, Alex Shao, Nathan Jew, Masayoshi Tomizuka \\
University of California, Berkeley \\
$^{*}$Equal contribution
}

\iclrfinalcopy % Uncomment for camera-ready version, but NOT for submission.
\begin{document}

\maketitle

% ============================================================
% Teaser
% ============================================================
\begin{figure}[h]
    \centering
    \includegraphics[
        width=\linewidth,
        keepaspectratio
    ]{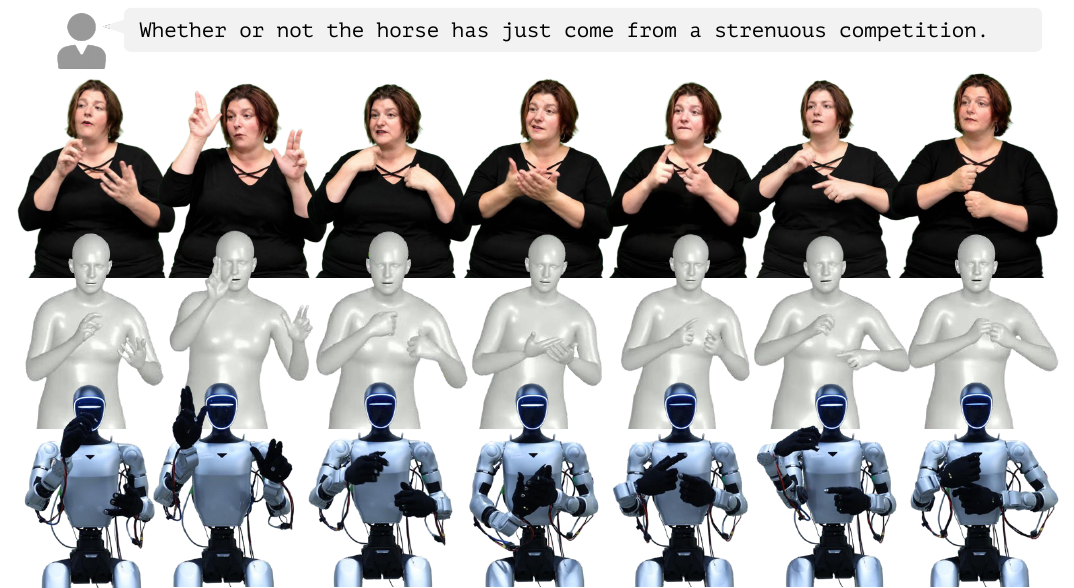}
    \caption{
        \textbf{From language to embodied robotic signing.}
        Given either speech or text, \ourmethod generates a continuous
        sign sequence and retargets it to a humanoid robot for
        physical execution.
    }
    \label{fig:teaser}
\end{figure}
\begin{abstract}
Sign-language interpretation in public communication relies on 
qualified professional interpreters and can be difficult to scale, motivating
robotic signing as a complementary accessibility interface. We present
\ourmethod, a text-conditioned sign language production (SLP) framework
for generating human-centric sign
motion that can be retargeted for robotic execution, with speech
supported optionally through an external ASR front end. Conventional autoregressive approaches flatten motion into a single
full-resolution token sequence, forcing long-range and local
dependencies to be modeled at a uniform temporal granularity.
\ourmethod instead combines part-wise Finite Scalar Quantization with
next-scale autoregression, generating motion over progressively finer
temporal resolutions while predicting synchronized body and hand tokens
in parallel within each step. This coarse-to-fine formulation provides
compact long-range context before progressively refining motion details,
while self-conditioning and context corruption improve robustness to
cross-scale prediction errors. The generated motion is subsequently
retargeted for physical humanoid execution. Extensive qualitative and quantitative evaluations are conducted to demonstrate the
effectiveness of \ourmethod.
\end{abstract}

\section{Introduction}

Sign languages are natural languages used by many deaf and
hard-of-hearing (DHH) people, conveying meaning through coordinated
handshapes, body motion, and non-manual markers such as facial
expressions and mouth movements. In public broadcasts, emergency
briefings, and press conferences, qualified professional interpreters
remain essential. Robotic signing is therefore not intended to
replace professional interpretation, but may serve as a complementary
accessibility channel for selected applications and resource-constrained
settings.

Recent work has advanced sign language production (SLP) from continuous
motion regression to discrete language-model-based generation, while
robotic systems have begun to explore embodied sign interaction
~\citep{signbot,signvla}. We develop a text-conditioned sign language motion generator that
produces human-centric body and bilateral-hand motion. The generated motion can then be transferred
to a humanoid robot through morphology-specific retargeting and control,
providing a practical path from SLP to physical execution. Speech is
supported optionally through a frozen off-the-shelf ASR (Automatic Speech Recognition) front end.
Although the human-centric generation includes facial and other
non-manual cues, the target robot lacks facial and lip actuation; these
cues are therefore omitted during embodiment, limiting the intelligibility
of signs that depend on non-manual markers. As illustrated in Fig.~\ref{fig:teaser}, \ourmethod translates sign motion
from language and retargets it to a humanoid robot for physical execution.

Existing sign language production (SLP) methods span both continuous
and discrete motion generation paradigms. Continuous regression and diffusion approaches directly model
continuous motion trajectories, while diffusion-based methods generate
motion through iterative denoising over multiple sampling steps~\citep{progressive_transformer,nar,nsa}.
Recent discrete approaches instead tokenize motion using VQ-based
representations~\citep{vqvae} and autoregress over flat temporal
sequences~\citep{t2sgpt,soke,m3t}. Such representations provide a natural
interface for pretrained language models~\citep{motiongpt,soke}, but
learned VQ codebooks can exhibit uneven utilization, while flat next-token decoding entangles long-range and local
dependencies within a single full-resolution autoregressive chain,
imposing a uniform temporal granularity throughout generation.

To address these limitations, we present \ourmethod, a hierarchical
autoregressive framework for embodied SLP. \ourmethod constructs
synchronized body and bilateral-hand token streams using part-wise
FSQ~\citep{fsq}. Unlike VQ methods that perform nearest-neighbor lookup in
a learned codebook, FSQ quantizes bounded scalar values over a fixed
grid, providing a regular discrete space for language modeling. On this
representation, an LM backbone~\citep{mt5} performs \emph{next-scale
autoregression}, inspired by coarse-to-fine generation
~\citep{var,moscale,infinitystar}. Coarser temporal scales
provide compact long-range context, while subsequent scales progressively
refine the motion at higher temporal resolution. Within each scale,
temporal positions remain autoregressive, while synchronized body and
bilateral-hand tokens are predicted in parallel.

Since errors in the coarse scale can be propagated through subsequent
refinement stages, we introduce self-conditioning (SC) and generic
context corruption (GCC). Both strategies expose the model to imperfect
preceding-scale contexts during training, reducing the train--test mismatch and improving robustness to errors propagated from coarser predictions.

Existing geometric metrics primarily measure spatial agreement with the
reference and can therefore under-characterize temporal
degeneration, such as near-static predictions or frozen terminal
segments. We introduce three complementary dynamics metrics---Dynamic
Degree, Magnitude Ratio, and Freeze-tail---to explicitly measure motion
activity, motion magnitude, and terminal freezing.

The generated human-centric motion is transferred to the target robot
through morphology-specific retargeting and tracking
~\citep{signbot,platform}, keeping sign generation independent of robot
morphology.

Overall, our contributions are as follows:
\begin{itemize}
    \item We propose a part-wise FSQ next-scale autoregressive model for
    sign language motion generation, jointly modeling body and bilateral-hand
    motion across progressively finer temporal resolutions. 
    \item We introduce self-conditioning (SC) and generic context
    corruption (GCC) to mitigate cross-scale exposure bias and improve
    robustness. 
    \item We introduce Dynamic Degree, Magnitude Ratio, and Freeze-tail
    to characterize degenerate motion dynamics that are not captured by
    conventional geometric or semantic metrics.
\end{itemize}

% We evaluate \ourmethod on the common How2Sign test subset.  Under a unified evaluation protocol, \ourmethod achieves state-of-the-art geometric performance; compared with the closely related SOKE baseline, it reduces body DTW-JPE by $32.5\%$.

%%%%%%%%%%%%%%%%%%%%%%%%%%%%%%%%%%%%%%%%%%%%%%%%%%%%%%%%%%%%%%%%%%%%%%%%%%%%%%%%
\section{Related Works}

\textbf{Sign Language Generation.}
Sign language generation (SLG) maps natural-language inputs to human signing motion. Early methods generate continuous pose sequences from text or gloss~\citep{progressive_transformer,adversarial_multichannel,nar,mcst_transformer,back_translation_3d,sign_characteristics}, later extending from sparse keypoints to holistic 3D avatars in both sentence- and word-level settings~\citep{nsa,signAvatars,wsigngen}. Recent work discretizes motion for language-model-based generation: MotionGPT~\citep{motiongpt} establishes this paradigm for general human motion, T2S-GPT~\citep{t2sgpt} applies autoregressive discrete modeling to SLG, SOKE~\citep{soke} combines part-wise body--hand tokenization with a pretrained multilingual LM and sign retrieval, and M$^3$T~\citep{m3t} further models body, hand, and facial motion with modality-specific tokens. We exclude M$^3$T from quantitative comparison because our robot lacks facial actuation and the code is not publicly released. 

Complementary work targets visual synthesis: SignLLM~\citep{signllm} generates pose sequences from text or prompts for subsequent video rendering, while other methods perform pose-conditioned~\citep{text2sign,signdiff} or end-to-end sign-video generation~\citep{signgen,stable_signer}. These works primarily target human sign visualization, making them less suitable for scenarios that require sign language to be physically performed on-site.

\textbf{Autoregressive and Next-Scale Motion Generation.}
Conventional next-token autoregression models motion through a single
full-resolution temporal sequence, entangling long-range and local
dependencies under a uniform temporal granularity. Next-scale autoregression instead generates global structure before local detail. VAR~\citep{var} introduces coarse-to-fine scale prediction for images; Infinity~\citep{infinity} scales this paradigm through bitwise prediction, and InfinityStar~\citep{infinitystar} extends it to unified spatio-temporal image and video generation. For human motion, MoScale~\citep{moscale} predicts progressively finer temporal resolutions so that global semantics are established before local dynamics. \ourmethod brings this principle to SLG by jointly modeling body and
hand streams across progressively finer temporal resolutions, using
coarser tokens for compact long-range context and finer
tokens for higher-resolution refinement.

\textbf{Embodied Robotic Signing.}
Existing robotic signing systems address related but distinct tasks. Human-to-robot tactile ASL transfer~\citep{platform} retargets human demonstrations to a bimanual robot, while SignVLA~\citep{signvla} maps finger-spelled signs to manipulation commands. SignBot~\citep{signbot} focuses on bidirectional human--robot sign interaction and largely integrates established components for sign understanding, response generation, sign generation, retargeting, and control. Its sign generator follows SOKE~\citep{soke}. In contrast, we formulate robotic signing as an embodied translation task and focus on advancing sign generation itself, introducing part-wise FSQ, next-scale autoregression, and robust cross-scale conditioning before robotic retargeting and execution.

\begin{figure*}[!bt]
    \centering
    \includegraphics[width=\textwidth]{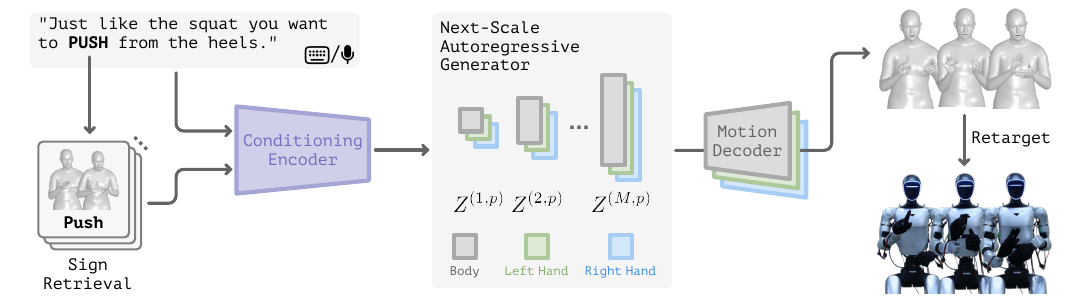}
    \caption{
        \textbf{Overview of \ourmethod.}
        Speech is first transcribed into text, while written input
        directly conditions the generator with retrieved signs.
        Part-wise FSQ tokenization represents sign motion at multiple
        temporal resolutions, which are generated sequentially through
        next-scale autoregression.
        The decoded human motion is subsequently retargeted to
        the robot in the real world.
    }
    \label{fig:overview_arch}
\end{figure*}

%%%%%%%%%%%%%%%%%%%%%%%%%%%%%%%%%%%%%%%%%%%%%%%%%%%%%%%%%%%%%%%%%%%%%%%%%%%%%%%%
\section{Method}
\label{sec:method}

% We present \textbf{\ourmethod},  a modular pipeline for robotic signing, as illustrated in Fig.~\ref{fig:overview_arch}. Speech inputs are first transcribed by a frozen, off-the-shelf ASR front end, whereas written inputs directly condition the generator. 

We present \textbf{\ourmethod} for robotic signing, as illustrated in
Fig.~\ref{fig:overview_arch}. The input can be either speech or text;
speech is first transcribed into text by a frozen, off-the-shelf ASR
front end. \ourmethod then translates the text into human-centric sign motion through next-scale autoregressive generation. The final output is the corresponding humanoid robot motion for sign-language translation.

% After language preprocessing, the pipeline comprises three stages: (i) part-wise FSQ tokenization of human-centric sign motion into synchronized multi-scale tokens; (ii) text-conditioned next-scale autoregressive generation from coarse to fine with robust cross-scale training; and (iii) embodiment transfer that reconstructs human motion and retargets it to the robot for low-level tracking.

\subsection{Part-Wise FSQ Motion Representation}
\label{sec:fsq}

\begin{figure}[t]
    \centering
    \includegraphics[width=\linewidth]{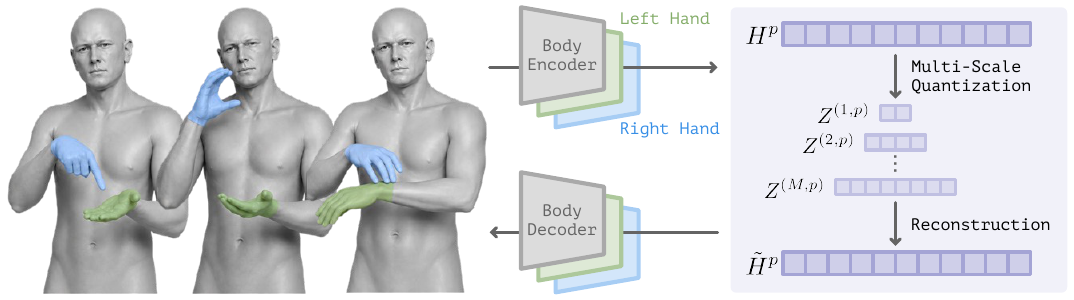}
    \caption{
        \textbf{Part-wise FSQ pyramid.}
        Finest-scale tokens are obtained from each motion stream, while
        coarser targets are constructed by temporal pooling and projection
        onto the FSQ grid.
        The hierarchy supervises next-scale generation, and only the finest
        scale is decoded for reconstruction.
    }
    \label{fig:fsq_pyramid}
\end{figure}

We represent each normalized human-centric sign sequence as
$
    S=[S^{B},S^{LH},S^{RH}]
    \in \mathbb{R}^{T\times d},
    \,
    d=d_{B}+d_{LH}+d_{RH},
$
where $T$ denotes the sequence length and
$S^{p}\in\mathbb{R}^{T\times d_p}$ is the motion stream for
$p\in\mathcal{P}=\{B,LH,RH\}$, corresponding to the body, left-hand,
and right-hand streams, respectively. The body stream additionally
includes facial motion parameters.
Because body and hand motions exhibit distinct statistics, a shared
tokenizer may underrepresent subtle manual articulation.
We therefore employ dedicated temporal encoders and decoders for each
part, as illustrated in Fig.~\ref{fig:fsq_pyramid}.

Given $S^p$, the corresponding encoder produces the finest-resolution
continuous latent sequence
\begin{equation}
    H^p=\mathcal{E}_p(S^p)
    \in\mathbb{R}^{L_M\times C}.
\end{equation}
Each latent vector is projected into a low-dimensional FSQ space,
$u_\tau^p=\mathcal{P}^{\mathrm{in}}_p(h_\tau^p)
\in\mathbb{R}^{d_q}$, and quantized independently along each scalar
dimension:
$
    \hat{u}_{\tau,j}^{p}
    =
    Q_{n_{p,j}}\!\left(u_{\tau,j}^{p}\right),
    \qquad
    K_p=\prod_{j=1}^{d_q}n_{p,j}.
$
The quantized tuple is converted through mixed-radix encoding into a
finest-scale token
$z_{\tau}^{(M,p)}\in\{1,\ldots,K_p\}$.
Gradients through the rounding are estimated using a
straight-through estimator.

Coarser token sequences are constructed directly from the finest-scale
tokens, without decoding and re-encoding motion.
We first dequantize the finest-scale IDs into their corresponding
$C$-dimensional implicit FSQ code vectors:
\begin{equation}
    E^{(M,p)}
    =
    \operatorname{Dequantize}_{p}
    \left(Z^{(M,p)}\right)
    \in\mathbb{R}^{L_M\times C}.
\end{equation}
For temporal divisors
$(d_1,\ldots,d_M)$, the target length of scale $m$ is
$
    L_m
    =
    \max\left(
        1,
        \left\lceil \frac{L_M}{d_m}\right\rceil
    \right).
$
For each scale $m<M$, the finest-scale code vectors are resized directly
to $L_m$ using one-dimensional adaptive average pooling and projected
back onto the fixed set of valid FSQ codes:
\begin{equation}
    \bar{E}^{(m,p)}
    =
    \operatorname{AdaptiveAvgPool}
    \left(E^{(M,p)},L_m\right),
    \quad
    z_{\tau}^{(m,p)}
    =
    \arg\min_{k\in\{1,\ldots,K_p\}}
    \left\|
        \bar{e}_{\tau}^{(m,p)}-c_k^p
    \right\|_2^2.
\end{equation}
where $\{c_k^p\}_{k=1}^{K_p}$ denotes the fixed set of dequantized FSQ
codes for part $p$. Adaptive pooling produces exactly $L_m$ output
positions for arbitrary $L_M$, without truncation or zero padding.
The resulting hierarchy is
$Z=\{Z^{(m,p)}\}_{m=1,\ldots,M;\,p\in\mathcal{P}}$, with
$L_1<\cdots<L_M$.
All coarser scales are independently derived from $Z^{(M,p)}$; the
construction is therefore neither recursive nor residual.
Exact pooling boundaries for non-divisible sequence lengths are given in
Appendix~\ref{app:multiscale_details}.

Unlike VQ-based tokenizers, finest-scale FSQ encoding does not require
nearest-neighbor lookup in a learned embedding codebook.
Its fixed Cartesian grid avoids learned-codebook utilization issues and
provides a regular prediction space.
The nearest-code operation above is used only to project pooled vectors
back onto the valid FSQ grid when constructing coarse-scale targets.
We adopt a compact quantization space to balance reconstruction fidelity
and downstream token predictability.

The complete hierarchy supervises coarse-to-fine generation, whereas
only the finest-scale tokens are decoded into motion.
For each part,
$
    \widetilde{H}^{p}
    =
    \operatorname{Dequantize}_{p}
    \left(Z^{(M,p)}\right),
    \quad
    \hat{S}^{p}
    =
    \mathcal{D}_{p}(\widetilde{H}^{p}),
$
and the reconstructed streams are concatenated as
$   \hat{S}
    =
    [\hat{S}^{B},\hat{S}^{LH},\hat{S}^{RH}].
$

\subsection{Next-Scale Sign Language Generation}
\label{sec:next_scale}

Let $x$ denote either a written input or the transcript produced by the
frozen ASR front end. \ourmethod employs a pretrained multilingual
language-model backbone to extract text features. Following
retrieval-enhanced SLP~\citep{soke}, we augment the language condition
with dictionary-based lexical sign retrieval. Input words are matched
against the dictionary in sentence order, and at most three unique
matches are retained. 

Their body and bilateral-hand FSQ prototypes are
serialized as lexical motion cues into the conditioning prompt, yielding
$C_x$. The complete sentence-level motion is still generated
autoregressively rather than assembled from retrieved sign clips.
Further details on dictionary construction, lexical matching, and
retrieval augmentation are provided in
Appendix~\ref{app:sign_retrieval}. A length predictor estimates the
finest-scale token length $L_M$, while coarser-scale lengths are
determined using fixed temporal ratios.

Rather than flattening all motion tokens into a single sequence, \ourmethod
factorizes generation causally across $M$ temporal scales:
\begin{equation}
\label{eq:next_scale_factorization}
\begin{split}
    p_\theta(Z\mid x)
    =
    \prod_{m=1}^{M}
    \prod_{\tau=1}^{L_m}
    \prod_{p\in\mathcal{P}}
    p_{\theta,p}\Big(
        z_{\tau}^{(m,p)}
        \,\big|\,
        C_x,\,
        Z^{(<m)},\,
        Z_{<\tau}^{(m)}
    \Big),
\end{split}
\end{equation}
where
$Z^{(<m)}=\{Z^{(j,p)}\mid j<m,\ p\in\mathcal{P}\}$
contains all previously generated coarser scales, and
$Z_{<\tau}^{(m)}=
\{z_{\tau'}^{(m,p)}\mid \tau'<\tau,\ p\in\mathcal{P}\}$
denotes the prefix of the current scale.
Here, next-scale autoregression refers to causal dependency across
temporal resolutions: scales are generated sequentially from coarse to
fine, while temporal positions remain autoregressive within each scale.
At each position, the shared decoder hidden state is projected onto
disjoint body, left-hand, and right-hand motion vocabularies, producing
the three synchronized tokens in parallel.

For scale $m>1$, let
$E_k\in\mathbb{R}^{L_k\times D}$ denote the motion-token embedding
sequence obtained from a preceding scale $k<m$.
We align it to the current length $L_m$ by deterministic zero-order
index resampling:
$
\label{eq:cross_scale_alignment}
    \left[
    \mathcal{A}_{k\rightarrow m}(E_k)
    \right]_t
    =
    (E_k)_{\left\lfloor tL_k/L_m\right\rfloor},
    \qquad
    t=0,\ldots,L_m-1.
$
Thus, coarse-scale embeddings are repeated over contiguous temporal
regions rather than linearly interpolated.

For each ordered scale pair $(m,k)$ with $k<m$, we learn a scalar gate
$
\label{eq:cross_scale_gate}
    g_{m,k}
    =
    \sigma(a_{m,k}),
$
where $a_{m,k}$ is a learned gate logit. The aligned preceding-scale
representations are fused into the cross-scale context
\begin{equation}
\label{eq:cross_scale_context}
    H_m
    =
    W_m
    \left(
    \frac{
        \sum_{k<m}
        g_{m,k}\,
        \mathcal{A}_{k\rightarrow m}(E_k)
    }{
        \sum_{k<m}g_{m,k}
    }
    \right),
\end{equation}
where $W_m$ is a learned scale-specific linear projection.
The resulting context is added to the current-scale decoder
representations. Standard causal self-attention restricts each position
to the current-scale prefix, while the language condition is accessed
through encoder--decoder cross-attention.
Further implementation details are provided in
Appendix~\ref{app:multiscale_details}.

Standard teacher forcing provides clean preceding-scale contexts during
training, whereas inference conditions finer scales on previously
predicted scales. To reduce this cross-scale exposure bias, we apply
self-conditioning (SC) followed by generic context corruption (GCC) to
preceding-scale contexts. For SC, each completed scale is replaced
sample-wise by its teacher-forced argmax prediction with probability
$p_{\mathrm{SC}}$; the replacement jointly covers body and bilateral-hand
tokens. GCC then independently replaces each valid context token with a
uniformly sampled code from the corresponding part vocabulary with
probability $p_{\mathrm{GCC}}$. Both probabilities follow linear warm-up schedules during training;
full specifications are provided in Appendix~\ref{app:sc_gcc}.

The generator is optimized with scale- and part-weighted token
prediction losses, together with auxiliary length prediction and
text--motion contrastive objectives:
\begin{equation}
\label{eq:generation_loss}
\mathcal{L}
=
\sum_{m=1}^{M}\lambda_m
\sum_{p\in\mathcal{P}}\alpha_p
\mathcal{L}_{\mathrm{CE}}^{(m,p)}
+
\lambda_{\mathrm{len}}\mathcal{L}_{\mathrm{len}}
+
\lambda_{\mathrm{con}}\mathcal{L}_{\mathrm{con}}.
\end{equation}
Here, $\mathcal{L}_{\mathrm{len}}$ supervises the finest-scale motion
length, while $\mathcal{L}_{\mathrm{con}}$ aligns text and motion
representations through a symmetric contrastive objective. Exact loss
definitions, weighting coefficients, and training hyperparameters are
provided in Appendix~\ref{app:training_objectives}. During inference,
\ourmethod generates scales sequentially from coarse to fine, after
which only the finest-scale tokens are decoded into continuous motion
$\hat{S}$.

\subsection{Embodied Retargeting and Tracking}
\label{sec:embodied_interface}

\ourmethod generates human-centric sign motion rather than robot joint
commands. From the decoded sequence, we recover SMPL-X upper-body
trajectories~\citep{smplx} and MANO hand articulation~\citep{mano}.
Following task-space human-to-robot transfer~\citep{signbot,platform},
the upper-body motion is mapped to the target morphology in a
torso-centered frame under joint-limit, reachability, self-collision,
and temporal-continuity constraints. Hand articulation is retargeted
independently from palm-normalized MANO geometry, and the resulting
body and hand references are tracked by a platform-specific low-level
controller. Motion channels unsupported by the target embodiment, including facial
expression parameters, are omitted during robotic execution. This modular interface keeps sign generation independent of
robot morphology; additional details on robotic retargeting and quantitative embodiment
evaluation are provided in Appendix~\ref{app:retargeting} and
Appendix~\ref{app:embodiment_eval}, respectively.

\section{Experiments}
\label{sec:experiments}

\begin{figure*}[t]
    \centering
    \includegraphics[width=0.95\textwidth]{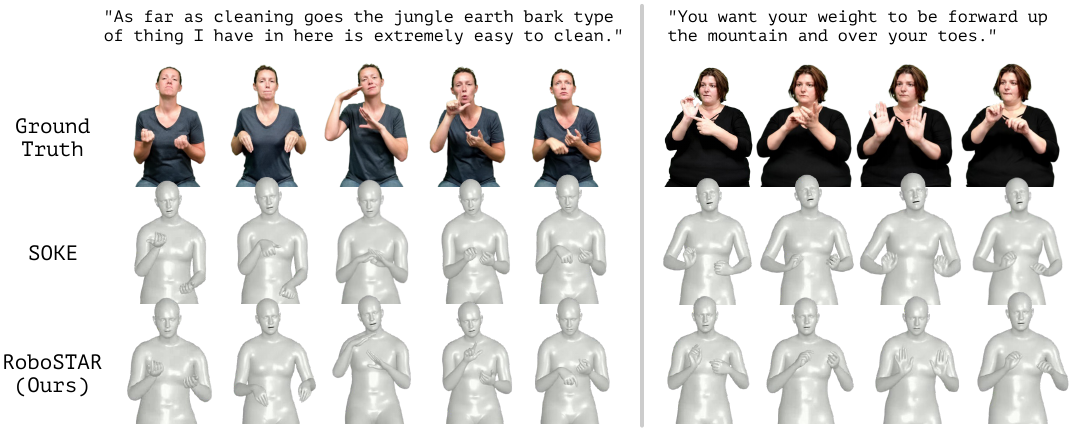}
    \vspace{-0.5cm}
    \caption{
        \textbf{Qualitative comparison on How2Sign.}
        Representative frames from ground-truth (GT), SOKE, and \ourmethod
        (Ours) sequences.
        \ourmethod better preserves the reference upper-body motion,
        hand articulation, and body--hand coordination.
    }
    \label{fig:qualitative_comparison}
\end{figure*}

\subsection{Experimental Setup}

%%%%%%%%%%%%%%%%%%%%%%%%%%%%%%%%%%%%%%%%%%%%%%%%%%%%%%%%%%%%%%%%%%%%%%%%%%%%%%%%
\begin{table*}[t]
    \centering
    \caption{
        Quantitative comparison on the common How2Sign test subset.
        GT is reported only as a reference and is excluded from
        ranking. B-$n$ denotes back-translation BLEU-$n$.
        Best and second-best results among the generation methods are shown
        in bold and underlined.
    }
    \label{tab:baseline_comparison}

    \renewcommand{\arraystretch}{1.15}
    \setlength{\tabcolsep}{1.2pt}
    \footnotesize

    \begin{tabular*}{\textwidth}{
        @{\extracolsep{\fill}}
        lccccccccccc
        @{}
    }
        \toprule

        \multirow{2}{*}{\textbf{Method}}
        & \multicolumn{2}{c}{\textbf{DTW-PA-JPE} $\downarrow$}
        & \multicolumn{2}{c}{\textbf{DTW-JPE} $\downarrow$}
        & \multicolumn{4}{c}{\textbf{B-T BLEU} $\uparrow$}
        & \multicolumn{3}{c}{\textbf{Motion Dynamics}} \\

        \cmidrule(lr){2-3}
        \cmidrule(lr){4-5}
        \cmidrule(lr){6-9}
        \cmidrule(lr){10-12}

        & \textbf{Body}
        & \textbf{Hand}
        & \textbf{Body}
        & \textbf{Hand}
        & \textbf{B-1}
        & \textbf{B-2}
        & \textbf{B-3}
        & \textbf{B-4}
        & \makecell{\textbf{Dynamic}\\$\leftrightarrow 0.8$}
        & \makecell{\textbf{Magnitude}\\$\leftrightarrow 1$}
        & \makecell{\textbf{Freeze-tail}\\$\downarrow$} \\

        \midrule

        GT
        & --
        & --
        & --
        & --
        & 22.45
        & 8.59
        & 3.95
        & 1.96
        & 0.804
        & 1.000
        & 0.036 \\

        \midrule

        PT
        & 10.85
        & 2.66
        & 10.84
        & 10.97
        & 13.44
        & 4.69
        & 1.99
        & 0.91
        & 0.245
        & 0.484
        & 0.232 \\

        NAR
        & \underline{4.21}
        & \underline{1.79}
        & \underline{4.14}
        & \underline{5.70}
        & 6.81
        & 2.10
        & 0.88
        & 0.42
        & 0.000
        & 0.004
        & 1.000 \\

        S-MotionGPT
        & 11.84
        & 3.92
        & 11.91
        & 13.30
        & 12.01
        & 4.33
        & 1.99
        & 1.03
        & 0.005
        & 0.016
        & 0.294 \\

        T2S-GPT
        & 4.35
        & 1.82
        & 4.27
        & 5.78
        & 7.13
        & 2.79
        & 1.28
        & 0.60
        & 0.000
        & 0.000
        & 1.000 \\

        SOKE
        & 5.73
        & 2.08
        & 5.73
        & 7.26
        & \underline{17.25}
        & \underline{6.31}
        & \underline{2.86}
        & \textbf{1.39}
        & \underline{0.388}
        & \underline{0.515}
        & \textbf{0.163} \\

        \midrule

        \textbf{Ours}
        & \textbf{3.96}
        & \textbf{1.52}
        & \textbf{3.87}
        & \textbf{5.34}
        & \textbf{20.08}
        & \textbf{7.12}
        & \textbf{3.04}
        & \underline{1.36}
        & \textbf{0.432}
        & \textbf{0.523}
        & \underline{0.187} \\

        \bottomrule
    \end{tabular*}
\end{table*}

\paragraph{Data Preparation.}
We evaluate \ourmethod on How2Sign~\citep{how2sign}, one of the largest
publicly available datasets for continuous ASL and a widely used
benchmark for sign language generation, containing approximately 35K
video--text pairs. We use the official training, validation, and test splits. For benchmark evaluation, the paired English transcripts are used as input to isolate motion generation from upstream speech-recognition errors. We reconstruct 3D motion using WHAC~\citep{whac} for the upper body and WiLoR~\citep{wilor} for MANO-compatible hand articulation. Each sequence is represented as $S\in\mathbb{R}^{T\times d}$, where $T$ is the number of frames and $d=133$ denotes the per-frame dimensionality, comprising body
and bilateral-hand pose parameters. We discard camera, lower-body, and identity-dependent shape that are irrelevant or unsupported by the target robot. 
% Retaining these channels or adopting higher-dimensional rotation representations enlarges the prediction space and destabilizes training.

\textbf{Evaluation Metrics.} Following prior 3D sign language production work~\citep{soke,m3t}, we
evaluate geometric fidelity using DTW-JPE and DTW-PA-JPE on native-length
sequences. DTW-JPE computes root- or wrist-aligned joint errors, whereas
DTW-PA-JPE additionally applies per-frame similarity Procrustes
alignment. We report cumulative DTW alignment cost in meters, averaging
the left- and right-hand errors.

To complement reference-based geometric metrics, we additionally measure
whether a generated sequence exhibits realistic temporal activity.
Dynamic Degree measures the fraction of sufficiently active transitions,
Magnitude Ratio compares the overall motion magnitude of generated and
ground-truth sequences, and Freeze-tail detects prolonged low-motion
segments at the end of a sequence. Dynamic Degree and Magnitude Ratio
have GT-derived reference values of approximately $0.8$ and $1.0$,
respectively, while lower Freeze-tail is preferred.

For semantic evaluation, we use a separately trained and frozen Hybrid
FSQ-T5 back-translator. Since the evaluator is itself imperfect, its BLEU
scores are interpreted comparatively rather than as an absolute measure
of translation accuracy. Full metric definitions, units, and
back-translation details are provided in
Appendix~\ref{app:evaluation_details}.

\textbf{Implementation Details.}
We train separate part-wise FSQ tokenizers with $4\times$ temporal
downsampling and initialize the generator from mT5-Large~\citep{mt5},
using four temporal divisors $(8,4,2,1)$. Speech, when provided, is
transcribed using the frozen Granite Speech 4.1 model
~\citep{granite-speech-4.1-2b}. Generated motions are retargeted to
Unitree G1 with dual 20-DoF Wuji hands. Full optimization and training
details are provided in Appendix~\ref{app:training_objectives}.

\subsection{Comparison with State-of-the-Art Methods}
\label{sec:main_results}

\textbf{Quantitative Results.}
We evaluate all available predictions on the common test subset using
the same joint topology, temporal alignment, and metric implementation.
As shown in Table~\ref{tab:baseline_comparison}, \ourmethod achieves the
lowest errors across all four geometric metrics, outperforming NAR, the
strongest prior baseline on geometric accuracy under our unified
evaluation protocol, on both body and hand motion. Notably, NAR achieves competitive geometric errors while exhibiting
near-zero Dynamic Degree and Magnitude Ratio together with a
Freeze-tail of $1.0$, revealing severely degenerate, near-static motion.
This contrast shows that favorable geometric errors alone can
mask motion collapse, motivating the complementary dynamics metrics to expose this failure mode.
\ourmethod also substantially improves all geometric metrics over the
closely related SOKE baseline.

For semantic fidelity, \ourmethod achieves the best B-1, B-2, and B-3
scores and the second-best B-4 score. It further produces the Dynamic
Degree and Magnitude Ratio closest to their reference targets, while
attaining the second-lowest Freeze-tail value. Together, these results
indicate that \ourmethod improves geometric and semantic fidelity while
maintaining substantially healthier motion dynamics.

\textbf{Qualitative Results.}
Figure~\ref{fig:qualitative_comparison} compares representative
sequences generated by \ourmethod and SOKE against the ground truth.
\ourmethod more closely follows the reference upper-body trajectories and
produces clearer hand articulation and body--hand coordination.

\begin{figure*}[t]
    \centering
    \includegraphics[width=0.9\textwidth]{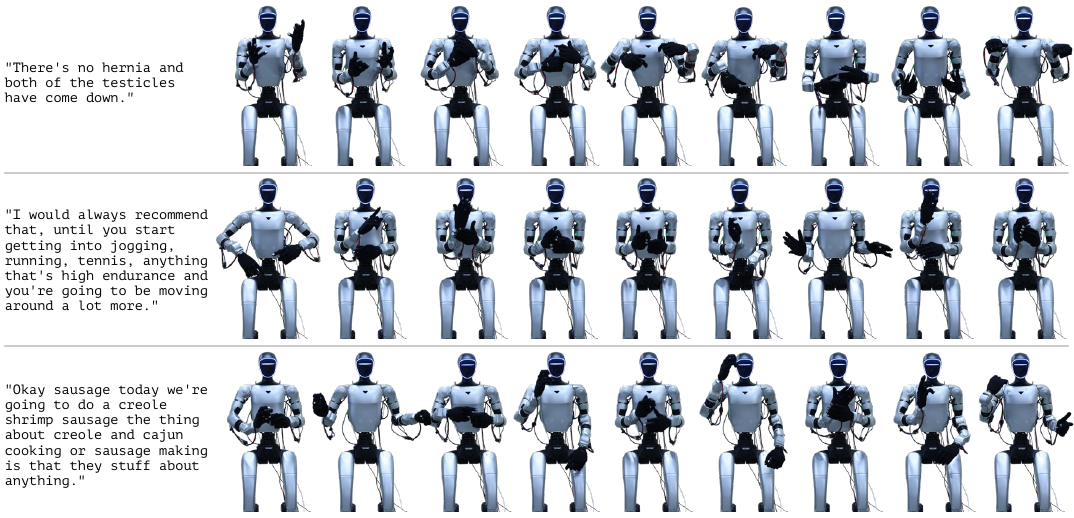}
    \caption{
        \textbf{Real-robot execution of \ourmethod-generated sign motion.}
        Representative frames show generated human-centric motions
        transferred to the physical robot through morphology-aware
        retargeting and low-level tracking.
    }
    \label{fig:real_robot}
\end{figure*}

\textbf{Real-Robot Execution.}
Figure~\ref{fig:real_robot} presents representative executions of
\ourmethod-generated motions after morphology-aware retargeting.
The physical robot reproduces coordinated upper-body and dexterous-hand
movements, demonstrating the feasibility of transferring the generated
human-centric motions to a robotic platform.

\textbf{Physical Embodiment Evaluation.}
We further quantify the distortion introduced by robotic embodiment.
Retargeting introduces measurable morphology-transfer error, while the
generated motions remain close to GT in positional feasibility, with
elbow and wrist errors of $2.0$\,cm and $5.6$\,cm for \ourmethod,
compared with $2.0$\,cm and $4.6$\,cm for GT motion.
Wrist orientation exhibits a larger residual, while closed-loop physics
simulation shows that the retargeted references remain trackable with a
$2.1$\,cm wrist error and $1.3^\circ$ joint-angle MAE.
Overall, these results indicate that the generated motions remain
reasonably compatible with the target robot morphology despite non-zero
retargeting error. Full definitions and results are provided in
Appendix~\ref{app:embodiment_eval}.

\subsection{Ablation Studies}
\label{sec:ablation}

\textbf{Effect of Next-Scale Generation.}
To isolate the contribution of next-scale autoregression, we compare
the complete four-scale hierarchy $\{8,4,2,1\}$ with a conventional
single-scale autoregressive model operating only at full temporal
resolution $\{1\}$. Both variants use the same backbone, tokenizer, and training protocol,
differing only in the temporal generation hierarchy.

\begin{wraptable}{r}{0.4\linewidth}
    \vspace{-6pt}
    \centering
    \caption{Effect of next-scale generation.}
    \label{tab:scale_ablation}

    \resizebox{\linewidth}{!}{
    \begin{tabular}{lcccc}
        \toprule
        \multirow{2}{*}{\textbf{Variant}}
        & \multicolumn{2}{c}{\textbf{DTW-PA-JPE} $\downarrow$}
        & \multicolumn{2}{c}{\textbf{DTW-JPE} $\downarrow$} \\
        \cmidrule(lr){2-3}
        \cmidrule(lr){4-5}
        & \textbf{Body}
        & \textbf{Hand}
        & \textbf{Body}
        & \textbf{Hand} \\
        \midrule

        Single-scale AR
        & 4.66
        & 2.56
        & 4.43
        & 6.81 \\

        \textbf{Next-scale}
        & \textbf{3.96}
        & \textbf{1.52}
        & \textbf{3.87}
        & \textbf{5.34} \\

        \bottomrule
    \end{tabular}
    }
    \vspace{-6pt}
\end{wraptable}

As shown in Table~\ref{tab:scale_ablation}, next-scale generation
consistently improves all geometric metrics. Compared with single-scale
autoregression, it reduces DTW-PA-JPE by $14.9\%$ for the body and
$40.4\%$ for the hands, and reduces DTW-JPE by $12.7\%$ and $21.5\%$,
respectively. These results demonstrate that coarse-to-fine temporal
generation provides a more effective formulation than flat
full-resolution autoregression, particularly for fine-grained hand
motion.

\textbf{Motion Tokenization.}
Table~\ref{tab:quantizer_ablation} compares FSQ and VQ-VAE under the
same part-wise architecture and a matched compression ratio of
$740.17\times$.
VQ-VAE achieves lower oracle reconstruction error, whereas FSQ
consistently yields better downstream generation across all geometric
metrics.
This result highlights a reconstruction--generation trade-off:
the tokenizer with higher reconstruction fidelity does not necessarily
provide the most predictable targets for autoregressive generation.
Despite its slightly weaker reconstruction, the regular FSQ space leads to more
accurate generated motions. Additional ablations on the language-model backbone and sign retrieval are provided in Appendix~\ref{app:additional_ablations}.

\begin{table}[t]
    \centering
    \caption{
        Reconstruction--generation trade-off between FSQ and VQ-VAE at a matched compression ratio, comparing tokenizer reconstruction fidelity and downstream motion generation.
    }
    \label{tab:quantizer_ablation}
    \vspace{5pt}
    \setlength{\tabcolsep}{2.2pt}
    \renewcommand{\arraystretch}{1.10}
    \scriptsize

    \begin{tabular}{lcccccc}
        \toprule

        \multirow{2}{*}{\textbf{Quantizer}}
        & \multicolumn{2}{c}{\textbf{Reconstruction}}
        & \multicolumn{4}{c}{\textbf{Generation}} \\

        \cmidrule(lr){2-3}
        \cmidrule(lr){4-7}

        & \textbf{Norm. MAE}
        & \textbf{Norm. RMSE}
        & \textbf{DTW-PA Body}
        & \textbf{DTW-PA Hand}
        & \textbf{DTW Body}
        & \textbf{DTW Hand} \\

        \midrule

        \textbf{FSQ}
        & 0.5221
        & 0.7053
        & \textbf{3.9643}
        & \textbf{1.5238}
        & \textbf{3.8672}
        & \textbf{5.3448} \\

        VQ-VAE
        & \textbf{0.4098}
        & \textbf{0.5569}
        & 4.7488
        & 2.1826
        & 4.6157
        & 6.3220 \\

        \bottomrule
    \end{tabular}
\end{table}

\begin{table*}[t]
    \centering
    \caption{
        Ablation study of self-conditioning (SC) and generic context
        corruption (GCC).
        The SC+GCC configuration is adopted as our final model.
        Best and second-best results are shown in bold and underlined.
    }
    \label{tab:sc_gcc_ablation}

    \renewcommand{\arraystretch}{1.15}
    \setlength{\tabcolsep}{2.0pt}
    \scriptsize

    \begin{tabular*}{\textwidth}{
        @{\extracolsep{\fill}}
        lccccccccccccc
        @{}
    }
        \toprule

        \multirow{2}{*}{\textbf{Configuration}}
        & \multicolumn{2}{c}{\textbf{Training}}
        & \multicolumn{2}{c}{\textbf{DTW-PA-JPE} $\downarrow$}
        & \multicolumn{2}{c}{\textbf{DTW-JPE} $\downarrow$}
        & \multicolumn{4}{c}{\textbf{B-T BLEU} $\uparrow$}
        & \multicolumn{3}{c}{\textbf{Motion Dynamics}} \\

        \cmidrule(lr){2-3}
        \cmidrule(lr){4-5}
        \cmidrule(lr){6-7}
        \cmidrule(lr){8-11}
        \cmidrule(lr){12-14}

        & \textbf{SC}
        & \textbf{GCC}
        & \textbf{Body}
        & \textbf{Hand}
        & \textbf{Body}
        & \textbf{Hand}
        & \textbf{B-1}
        & \textbf{B-2}
        & \textbf{B-3}
        & \textbf{B-4}
        & \makecell{\textbf{Dynamic}\\$\leftrightarrow 0.8$}
        & \makecell{\textbf{Magnitude}\\$\leftrightarrow 1$}
        & \makecell{\textbf{Freeze-tail}\\$\downarrow$} \\

        \midrule

        Neither
        & \xmark
        & \xmark
        & 4.19
        & 1.85
        & 4.06
        & 5.90
        & 8.85
        & 3.02
        & 1.37
        & 0.63
        & 0.13
        & 0.23
        & \textbf{0.16} \\

        GCC only
        & \xmark
        & \cmark
        & 4.25
        & 1.96
        & 4.10
        & 6.04
        & 8.17
        & 2.74
        & 1.23
        & 0.55
        & 0.15
        & 0.25
        & 0.23 \\

        SC only
        & \cmark
        & \xmark
        & \textbf{3.92}
        & \underline{1.54}
        & \textbf{3.82}
        & \textbf{5.07}
        & \underline{17.93}
        & \underline{6.34}
        & \underline{2.67}
        & \underline{1.14}
        & \underline{0.34}
        & \underline{0.47}
        & 0.31 \\

        \midrule

        \textbf{Ours (SC+GCC)}
        & \cmark
        & \cmark
        & \underline{3.96}
        & \textbf{1.52}
        & \underline{3.87}
        & \underline{5.34}
        & \textbf{20.08}
        & \textbf{7.12}
        & \textbf{3.04}
        & \textbf{1.36}
        & \textbf{0.43}
        & \textbf{0.52}
        & \underline{0.19} \\

        \bottomrule
    \end{tabular*}
\end{table*}

\textbf{Self-conditioning and context corruption.}
Table~\ref{tab:sc_gcc_ablation} studies self-conditioning (SC) and
generic context corruption (GCC). GCC alone degrades geometric
accuracy, whereas SC substantially improves it by exposing finer scales
to model-generated histories. Adding GCC to SC introduces a modest
geometric trade-off but substantially improves back-translation and
motion-dynamics metrics. We therefore
adopt SC+GCC as the final configuration, balancing geometric fidelity,
semantic consistency, and motion robustness.

\subsection{User Study}
\label{sec:user_study}

\begin{wrapfigure}{r}{0.43\linewidth}
    \vspace{-12pt}
    \centering
    \includegraphics[width=\linewidth]{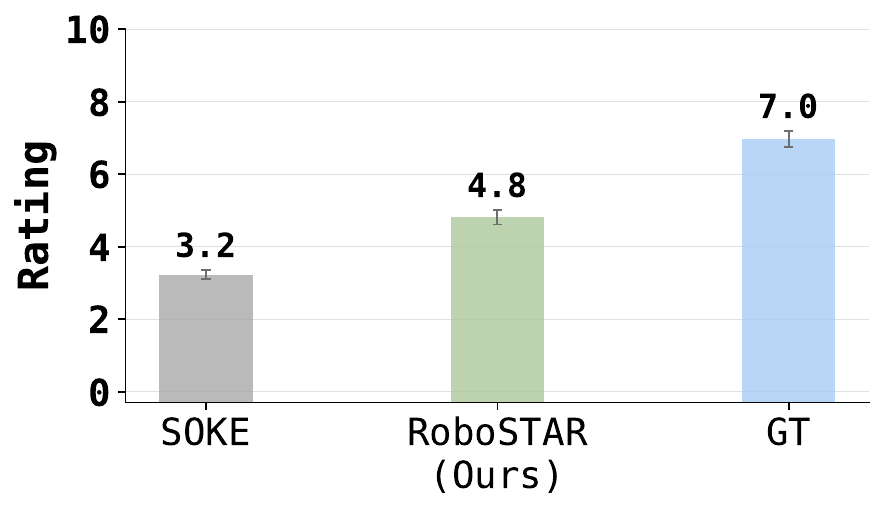}
    \vspace{-0.8cm}
    \caption{
        \textbf{User study.}
        Mean ratings; error bars denote SEM.
    }
    \label{fig:user}
    % \vspace{-10pt}
\end{wrapfigure}

We conduct a blinded user study with four fluent signers using 15
held-out sentences from the How2Sign dataset. For each sentence,
participants view the ground-truth motion, SOKE output, and \ourmethod
output in randomized order with method identities hidden. Each motion
receives a single overall score on a 1--10 scale, jointly considering
semantic accuracy, hand-shape clarity, motion naturalness, and temporal
fluency. Ratings are first averaged across the four participants for
each sentence, and we report the mean and standard error of the mean
(SEM) across the resulting 15 sentence-level scores.

As shown in Fig.~\ref{fig:user}, \ourmethod receives higher mean ratings
than SOKE and narrows the perceptual gap to the ground-truth motions.
Together, these ratings provide complementary evidence that \ourmethod
is a promising sign-generation backbone for downstream applications,
including public information delivery and embodied accessibility systems.

%%%%%%%%%%%%%%%%%%%%%%%%%%%%%%%%%%%%%%%%%%%%%%%%%%%%%%%%%%%%%%%%%%%%%%%%%%%%%%%%
\section{Conclusion}

In this paper, we introduced \ourmethod, a framework for 
robotic sign language translation.
At its core, \ourmethod combines part-wise FSQ with next-scale
autoregression to generate synchronized body and hand motion across
progressively finer temporal resolutions, while self-conditioning and
generic context corruption improve robustness to cross-scale prediction
errors. We further introduce three complementary motion-dynamics metrics
to characterize failure modes that are not captured by conventional
geometric or semantic evaluation. Under a unified evaluation protocol on
reconstructed 3D motion from the common How2Sign test subset, \ourmethod
achieves state-of-the-art geometric performance.
Physical robot demonstrations further illustrate the feasibility of
executing the generated motions on a humanoid platform, promising
future embodied accessibility applications for DHH communities.

\textbf{Limitations and Future Work.}
Our system is currently evaluated only on How2Sign for ASL, leaving cross-domain and cross-language generalization for future study. Training motions are reconstructed from monocular videos, making fine-grained hand articulation and occlusions sensitive to 3D reconstruction errors. In addition, the current robot lacks facial and lip actuation for non-manual markers. Future work will broaden data and language coverage, improve 3D supervision, and explore richer robotic embodiments.

% \textbf{Limitations and Future Work}
% Our current system has several limitations. First, it is developed and
% evaluated only on How2Sign for American Sign Language (ASL), and may
% inherit dataset- and signer-specific biases that limit generalization to
% unseen domains and other sign languages. RWTH-PHOENIX-Weather-2014T,
% which contains German Sign Language (DGS), and CSL-Daily, which contains
% Chinese Sign Language (CSL), are not included in the current study due
% to dataset-access and application constraints. We encourage future work
% to transfer and evaluate our framework across additional sign languages
% and datasets. Second, the training motions are
% reconstructed from monocular videos, so 3D reconstruction errors,
% particularly in fine-grained hand articulation and occluded poses, may
% propagate into the learned representations. Third, although our system
% accepts speech through an off-the-shelf ASR front end, our main evaluation
% uses ground-truth transcripts and therefore does not assess error
% propagation in the complete speech-to-sign pipeline. Finally, the current
% robot cannot reproduce facial expressions or lip movements, limiting
% signs that rely on facial markers. Future work will address these
% limitations through more diverse data, improved 3D supervision,
% end-to-end speech evaluation, and richer robotic embodiments.

\clearpage
\subsection*{AI Use Statement}

% 不计入 9 页 page limit

In this work, generative AI tools were used in two limited capacities.
First, AI-based enhancement was applied to improve the visual clarity
of blurred source videos for visualization purposes only; the enhanced
videos were not used to produce the reported quantitative results. Second, generative AI was used for language
polishing and editorial assistance during manuscript preparation.
All AI-assisted outputs were reviewed and, where necessary, revised by us. Generative AI was not used to generate experimental
results, quantitative evaluations, human-study responses, or scientific
conclusions. We take responsibility for the final content of this work, including text, claims or artifacts produced with the aid of generative AI.

\subsection*{Ethics Statement}

This work explores robotic signing as a complementary accessibility
technology rather than a replacement for qualified professional sign
language interpreters, particularly in emergency, medical, legal, or
other high-stakes settings where mistranslation may have serious
consequences. The current system is evaluated on ASL data from How2Sign
and may inherit dataset- and signer-specific biases. Moreover, the target
robot cannot reproduce facial and lip-based non-manual markers, which
may reduce intelligibility for signs that depend on these cues. These
limitations should be considered before any real-world deployment.

Our user study involved four fluent signers. All participants were
informed about the study procedure and the intended research use of their
responses, and voluntarily consented to participate. No
participant-identifying information is reported.

We use the system only for research evaluation and do not claim that it is
ready for unsupervised deployment in safety-critical communication.

\subsection*{Reproducibility Statement}

% 不计入 9 页 page limit

We provide implementation and evaluation details throughout the main
paper and appendix to facilitate reproducibility. The data processing,
model configuration, and main experimental protocol are described in
Sec.~\ref{sec:experiments}.  Additional implementation details for multi-scale construction and
cross-scale conditioning, sign retrieval, self-conditioning and context
corruption, the training objectives, and robotic retargeting are provided
in Appendix~\ref{app:multiscale_details},
Appendix~\ref{app:sign_retrieval},
Appendix~\ref{app:sc_gcc},
Appendix~\ref{app:training_objectives}, and
Appendix~\ref{app:retargeting}, respectively. Complete definitions and units for the
geometric and motion-dynamics metrics, together with the architecture,
data splits, and held-out performance of the back-translation evaluator,
are provided in Appendix~\ref{app:evaluation_details}.

\bibliography{iclr2027_conference}

@String(CVPR  = {Conference on Computer Vision and Pattern Recognition})

@String(ICCV  = {International Conference on Computer Vision})

@String(NeurIPS  = {Conference and Workshop on Neural Information Processing Systems})

@String(ICLR  = {International Conference on Learning Representations})

@String(ECCV  = {European Conference on Computer Vision})

@String(BMVC = {British Machine Vision Conference})

@String(ACL = {Annual Meeting of the Association for Computational Linguistics})

@String(ICRA = {International Conference on Robotics and Automation})

@inproceedings{soke,
  author       = {Ronglai Zuo and
                  Rolandos Alexandros Potamias and
                  Evangelos Ververas and
                  Jiankang Deng and
                  Stefanos Zafeiriou},
  title        = {Signs as Tokens: {A} Retrieval-Enhanced Multilingual Sign Language
                  Generator},
  booktitle    = ICCV,
  year         = {2025},
}

@inproceedings{how2sign,
  author       = {Amanda Cardoso Duarte and
                  Shruti Palaskar and
                  Lucas Ventura and
                  Deepti Ghadiyaram and
                  Kenneth DeHaan and
                  Florian Metze and
                  Jordi Torres and
                  Xavier Gir{\'{o}}{-}i{-}Nieto},
  title        = {How2Sign: {A} Large-Scale Multimodal Dataset for Continuous American
                  Sign Language},
  booktitle    = CVPR,
  year         = {2021},
}

@inproceedings{progressive_transformer,
  author       = {Ben Saunders and
                  Necati Cihan Camg{\"{o}}z and
                  Richard Bowden},
  editor       = {Andrea Vedaldi and
                  Horst Bischof and
                  Thomas Brox and
                  Jan{-}Michael Frahm},
  title        = {Progressive Transformers for End-to-End Sign Language Production},
  booktitle    = ECCV,
  year         = {2020},
}

@inproceedings{nar,
  author       = {Eui Jun Hwang and
                  Jung{-}Ho Kim and
                  Jong C. Park},
  title        = {Non-Autoregressive Sign Language Production with Gaussian Space},
  booktitle    = BMVC,
  year         = {2021},
}

@inproceedings{nsa,
  author       = {Vasileios Baltatzis and
                  Rolandos Alexandros Potamias and
                  Evangelos Ververas and
                  Guanxiong Sun and
                  Jiankang Deng and
                  Stefanos Zafeiriou},
  title        = {Neural Sign Actors: {A} diffusion model for 3D sign language production
                  from text},
  booktitle    = CVPR,
  year         = {2024},
}

@inproceedings{t2sgpt,
  author       = {Aoxiong Yin and
                  Haoyuan Li and
                  Kai Shen and
                  Siliang Tang and
                  Yueting Zhuang},
  title        = {{T2S-GPT:} Dynamic Vector Quantization for Autoregressive Sign Language
                  Production from Text},
  booktitle    = ACL,
  year         = {2024},
}

@inproceedings{motiongpt,
  author       = {Biao Jiang and
                  Xin Chen and
                  Wen Liu and
                  Jingyi Yu and
                  Gang Yu and
                  Tao Chen},
  title        = {MotionGPT: Human Motion as a Foreign Language},
  booktitle    = NeurIPS,
  year         = {2023},
}

@article{m3t,
  author       = {Alexandre Symeonidis{-}Herzig and
                  JianHe Low and
                  Ozge Mercanoglu Sincan and
                  Richard Bowden},
  title        = {{M3T:} Discrete Multi-Modal Motion Tokens for Sign Language Production},
  journal      = {arXiv preprint},
  year         = {2026},
}

@inproceedings{whac,
  author       = {Wanqi Yin and
                  Zhongang Cai and
                  Ruisi Wang and
                  Fanzhou Wang and
                  Chen Wei and
                  Haiyi Mei and
                  Weiye Xiao and
                  Zhitao Yang and
                  Qingping Sun and
                  Atsushi Yamashita and
                  Ziwei Liu and
                  Lei Yang},
  title        = {{WHAC:} World-Grounded Humans and Cameras},
  booktitle    = ECCV,
  year         = {2024},
}

@inproceedings{wilor,
  author       = {Rolandos Alexandros Potamias and
                  Jinglei Zhang and
                  Jiankang Deng and
                  Stefanos Zafeiriou},
  title        = {WiLoR: End-to-end 3D Hand Localization and Reconstruction in-the-wild},
  booktitle    = CVPR,
  year         = {2025},
}

@inproceedings{smplx,
  author       = {Georgios Pavlakos and
                  Vasileios Choutas and
                  Nima Ghorbani and
                  Timo Bolkart and
                  Ahmed A. A. Osman and
                  Dimitrios Tzionas and
                  Michael J. Black},
  title        = {Expressive Body Capture: 3D Hands, Face, and Body From a Single Image},
  booktitle    = CVPR,
  year         = {2019},
}

@article{mano,
  author       = {Javier Romero and
                  Dimitrios Tzionas and
                  Michael J. Black},
  title        = {Embodied hands: modeling and capturing hands and bodies together},
  journal      = {{ACM} Trans. Graph.},
  year         = {2017},
}

@inproceedings{vqvae,
  author       = {A{\"{a}}ron van den Oord and
                  Oriol Vinyals and
                  Koray Kavukcuoglu},
  title        = {Neural Discrete Representation Learning},
  booktitle    = NeurIPS,
  year         = {2017},
}

@inproceedings{fsq,
  author       = {Fabian Mentzer and
                  David Minnen and
                  Eirikur Agustsson and
                  Michael Tschannen},
  title        = {Finite Scalar Quantization: {VQ-VAE} Made Simple},
  booktitle    = ICLR,
  year         = {2024},
}

@inproceedings{mt5,
  author       = {Linting Xue and
                  Noah Constant and
                  Adam Roberts and
                  Mihir Kale and
                  Rami Al{-}Rfou and
                  Aditya Siddhant and
                  Aditya Barua and
                  Colin Raffel},
  title        = {mT5: {A} Massively Multilingual Pre-trained Text-to-Text Transformer},
  booktitle    = {Proceedings of the 2021 Conference of the North American Chapter of
                  the Association for Computational Linguistics: Human Language Technologies,
                  {NAACL-HLT} 2021, Online, June 6-11, 2021},
  year         = {2021},
}

@inproceedings{var,
  author       = {Keyu Tian and
                  Yi Jiang and
                  Zehuan Yuan and
                  Bingyue Peng and
                  Liwei Wang},
  title        = {Visual Autoregressive Modeling: Scalable Image Generation via Next-Scale
                  Prediction},
  booktitle    = NeurIPS,
  year         = {2024},
}

@inproceedings{moscale,
  author       = {Zhiwei Zheng and
                  Shibo Jin and
                  Lingjie Liu and
                  Mingmin Zhao},
  title        = {Next-Scale Autoregressive Models for Text-to-Motion Generation},
  booktitle    = CVPR,
  year         = {2026},
}

@inproceedings{infinitystar,
  author       = {Jinlai Liu and
                  Jian Han and
                  Bin Yan and
                  Hui Wu and
                  Fengda Zhu and
                  Xing Wang and
                  Yi Jiang and
                  Bingyue Peng and
                  Zehuan Yuan},
  title        = {InfinityStar: Unified Spacetime AutoRegressive Modeling for Visual
                  Generation},
  booktitle    = NeurIPS,
  year         = {2025},
}

@inproceedings{signbot,
  author       = {Guanren Qiao and
                  Sixu Lin and
                  Ronglai Zuo and
                  Zhizheng Wu and
                  Kui Jia and
                  Guiliang Liu},
  title        = {SignBot: Learning Human-to-Humanoid Sign Language Interaction},
  booktitle    = ICRA,
  year         = {2026},
}

@inproceedings{signAvatars,
  author       = {Zhengdi Yu and
                  Shaoli Huang and
                  Yongkang Cheng and
                  Tolga Birdal},
  title        = {SignAvatars: {A} Large-Scale 3D Sign Language Holistic Motion Dataset
                  and Benchmark},
  booktitle    = ECCV,
  year         = {2024},
}

@inproceedings{infinity,
  author       = {Jian Han and
                  Jinlai Liu and
                  Yi Jiang and
                  Bin Yan and
                  Yuqi Zhang and
                  Zehuan Yuan and
                  Bingyue Peng and
                  Xiaobing Liu},
  title        = {Infinity: Scaling Bitwise AutoRegressive Modeling for High-Resolution
                  Image Synthesis},
  booktitle    = CVPR,
  year         = {2025},
}

@inproceedings{wsigngen,
  author       = {Lu Dong and
                  Xiao Wang and
                  Ifeoma Nwogu},
  title        = {Word-Conditioned 3D American Sign Language Motion Generation},
  booktitle    = {Findings of the Association for Computational Linguistics: {EMNLP}
                  },
  year         = {2024},
}

@article{signvla,
  author       = {Xinyu Tan and
                  Ningwei Bai and
                  Harry Gardener and
                  Zhengyang Zhong and
                  Luoyu Zhang and
                  Liuhaichen Yang and
                  Zhekai Duan and
                  Monkgogi Galeitsiwe and
                  Zezhi Tang},
  title        = {SignVLA: {A} Gloss-Free Vision-Language-Action Framework for Real-Time
                  Sign Language-Guided Robotic Manipulation},
  journal      = {arXiv},
  year         = {2026},
}

@inproceedings{platform,
  author       = {Che{-}Ming Chang and
                  Felipe Sanches and
                  Geng Gao and
                  Minas Liarokapis},
  title        = {On Human to Robot Skill Transfer for the Execution of Complex Tactile
                  American Sign Language Tasks with a Bimanual Robot Platform},
  booktitle    = {46th Annual International Conference of the {IEEE} Engineering in
                  Medicine and Biology Society, {EMBC}},
  year         = {2024},
}

@inproceedings{stable_signer,
  author       = {Sen Fang and
                  Yalin Feng and
                  Hongbin Zhong and
                  Yanxin Zhang and
                  Dimitris N. Metaxas},
  title        = {Stable Signer: Hierarchical Sign Language Generative Model},
  booktitle    = ACL,
  year         = {2026},
}

@inproceedings{signllm,
  author       = {Sen Fang and
                  Chen Chen and
                  Lei Wang and
                  Ce Zheng and
                  Chunyu Sui and
                  Yapeng Tian},
  title        = {SignLLM: Sign Language Production Large Language Models},
  booktitle    = {{IEEE/CVF} International Conference on Computer Vision
                  - Workshops},
  year         = {2025},
}

@article{sign_characteristics,
  author       = {Manuel Gil{-}Mart{\'{\i}}n and
                  Mar{\'{\i}}a Villa{-}Monedero and
                  Andrzej Pomirski and
                  Daniel S{\'{a}}ez{-}Trigueros and
                  Rub{\'{e}}n San Segundo},
  title        = {Sign Language Motion Generation from Sign Characteristics},
  journal      = {Sensors},
  year         = {2023},
}

@inproceedings{mcst_transformer,
  author       = {Xiaohan Ma and
                  Rize Jin and
                  Tae{-}Sun Chung},
  title        = {Multi-Channel Spatio-Temporal Transformer for Sign Language Production},
  booktitle    = {Proceedings of the 2024 Joint International Conference on Computational
                  Linguistics, Language Resources and Evaluation, {LREC/COLING}},
  year         = {2024},
}

@inproceedings{adversarial_multichannel,
  author       = {Ben Saunders and
                  Richard Bowden and
                  Necati Cihan Camg{\"{o}}z},
  title        = {Adversarial Training for Multi-Channel Sign Language Production},
  booktitle    = {31st British Machine Vision Conference 2020, {BMVC}},
  year         = {2020},
}

@inproceedings{back_translation_3d,
  author       = {Stephanie Stoll and
                  Armin Mustafa and
                  Jean{-}Yves Guillemaut},
  title        = {There and Back Again: 3D Sign Language Generation from Text Using
                  Back-Translation},
  booktitle    = {International Conference on 3D Vision, 3DV},
  year         = {2022},
}

@ARTICLE{dtw,
  author={Sakoe, H. and Chiba, S.},
  journal={IEEE Transactions on Acoustics, Speech, and Signal Processing}, 
  title={Dynamic programming algorithm optimization for spoken word recognition}, 
  year={1978},}

@misc{granite-speech-4.1-2b,
  title={Granite 4.1 Speech},
  author={IBM Granite Speech Team},
  year={2026},
  url={https://huggingface.co/ibm-granite/granite-speech-4.1-2b}
}

@inproceedings{signdiff,
  author       = {Sen Fang and
                  Chunyu Sui and
                  Yanghao Zhou and
                  Xuedong Zhang and
                  Hongbin Zhong and
                  Yapeng Tian and
                  Chen Chen},
  title        = {SignDiff: Diffusion Model for American Sign Language Production},
  booktitle    = {19th {IEEE} International Conference on Automatic Face and Gesture
                  Recognition, {FG}},
  year         = {2025},
}

@article{text2sign,
  author       = {Stephanie Stoll and
                  Necati Cihan Camg{\"{o}}z and
                  Simon Hadfield and
                  Richard Bowden},
  title        = {Text2Sign: Towards Sign Language Production Using Neural Machine Translation
                  and Generative Adversarial Networks},
  journal      = {Int. J. Comput. Vis.},
  year         = {2020},
}

@inproceedings{signgen,
  author       = {Fan Qi and
                  Yu Duan and
                  Huaiwen Zhang and
                  Changsheng Xu},
  title        = {SignGen: End-to-End Sign Language Video Generation with Latent Diffusion},
  booktitle    = ECCV,
  year         = {2024},
}
\bibliographystyle{iclr2027_conference}

\clearpage

\appendix
\section{Appendix}

% ============================================================
% A.1 Multi-Scale Construction and Cross-Scale Conditioning
% ============================================================
\subsection{Multi-Scale Construction and Cross-Scale Conditioning}
\label{app:multiscale_details}

\paragraph{Temporal pyramid construction.}
Let $L=L_M$ denote the finest-scale FSQ sequence length.
For temporal divisor $d_m\in\{8,4,2,1\}$, the target scale length is
\begin{equation}
    L_m
    =
    \max\left(
        1,
        \left\lceil \frac{L}{d_m}\right\rceil
    \right).
\end{equation}
The body, left-hand, and right-hand streams are processed independently
using their corresponding FSQ code vectors, while sharing the same
target length $L_m$. Every coarser scale is constructed directly from
the finest-scale sequence rather than recursively from the immediately
finer scale.

For output position $r=0,\ldots,L_m-1$, adaptive average pooling uses
the interval
\begin{equation}
    a_r
    =
    \left\lfloor
        \frac{rL}{L_m}
    \right\rfloor,
    \qquad
    b_r
    =
    \left\lceil
        \frac{(r+1)L}{L_m}
    \right\rceil,
\end{equation}
and computes
\begin{equation}
    \bar{\mathbf q}^{(m)}_r
    =
    \frac{1}{b_r-a_r}
    \sum_{i=a_r}^{b_r-1}
    \mathbf q_i ,
\end{equation}
where $\mathbf q_i$ is the dequantized FSQ code vector at finest-scale
position $i$. Each pooled vector is subsequently assigned to its nearest
valid FSQ code. Hence, when $L$ is not divisible by the temporal
divisor, no tail tokens are discarded and no zero padding is introduced.
For $d_m=1$, the finest-scale token sequence is copied directly without
pooling or requantization.

\paragraph{Cross-scale alignment and gating.}
Cross-scale context alignment is distinct from the adaptive pooling used
to construct supervision targets. For a preceding scale $k<m$, target
position $t$ at scale $m$ gathers source position
\begin{equation}
    j(t)
    =
    \left\lfloor
        \frac{tL_k}{L_m}
    \right\rfloor,
\end{equation}
such that
\begin{equation}
    \left[
    \mathcal A_{k\rightarrow m}(E_k)
    \right]_t
    =
    (E_k)_{j(t)} .
\end{equation}

The gate logits form a learned matrix
$A\in\mathbb{R}^{M\times M}$. For $k<m$,
\begin{equation}
    g_{m,k}
    =
    \sigma(A_{m,k}),
\end{equation}
and the cross-scale context is
\begin{equation}
    H_m
    =
    W_m
    \left(
    \frac{
        \sum_{k<m}
        g_{m,k}\,
        \mathcal A_{k\rightarrow m}(E_k)
    }{
        \sum_{k<m}g_{m,k}
    }
    \right).
\end{equation}
Here each $g_{m,k}$ is a single scalar shared across samples, temporal
positions, and embedding dimensions, and $W_m$ is a scale-specific
bias-free linear projection. The gate logits are initialized to zero,
so $g_{m,k}=0.5$ initially and all available preceding scales contribute
equally after normalization.

For the coarsest scale $m=1$, where no preceding scale is available, a
learned seed vector $\mathbf{s}_1$ is repeated over temporal positions
and projected as
\begin{equation}
    H_1=W_1\mathbf{s}_1.
\end{equation}

% ============================================================
% A.2 Sign Retrieval
% ============================================================
\subsection{Sign Retrieval}
\label{app:sign_retrieval}

We construct the lexical memory from the isolated-word training split of
\texttt{akasheroor/American-Sign-Language-Dataset}, resulting in 2,206
dictionary entries. For each lexical item, all available training
instances are encoded using the same part-wise FSQ tokenizer as the main
model, and the instance with the lowest tokenizer reconstruction MAE is
retained as its prototype. Each body, left-hand, and right-hand code
sequence is deterministically normalized to 10 FSQ tokens.

Retrieval is implemented as lexical dictionary lookup rather than
embedding-based nearest-neighbor search. Given an input sentence, words
are normalized to uppercase alphanumeric forms, with a lightweight
suffix-based lemma used as fallback. Exact dictionary matches are
collected in their original sentence order, duplicate entries are
removed, and the first three matches are retained:
\begin{equation}
R(x)
=
\operatorname{FirstK}
\left(
\{q_i \mid q_i \in \mathcal D\}
\right),
\qquad K=3.
\end{equation}
Thus, $0\leq |R(x)|\leq3$. Each retrieved prototype contributes 10
body, 10 left-hand, and 10 right-hand FSQ codes, for at most 90 raw
retrieved code IDs per sentence. These codes are serialized into the
textual mT5 conditioning prompt and are used only as auxiliary lexical
motion cues; they are not concatenated to the generated motion.

During training, retrieval augmentation is independently dropped for
each sample with probability $0.10$ to reduce dependence on dictionary
coverage. Validation and inference use retrieval without this dropout.

% ============================================================
% A.2 Self-Conditioning and Context Corruption
% ============================================================
\subsection{Self-Conditioning and Context Corruption}
\label{app:sc_gcc}

SC and GCC modify only preceding-scale contexts during training and do
not alter the current-scale supervision. After the teacher-forced
forward pass at scale $m$, we obtain deterministic part-wise predictions
by restricted-vocabulary argmax:
\begin{equation}
\hat z_{\tau}^{(m,p)}
=
\arg\max_{v\in\mathcal V_p}
\ell_{\tau,v}^{(m,p)},
\end{equation}
with vocabulary sizes
$|\mathcal V_B|=96$ and
$|\mathcal V_{LH}|=|\mathcal V_{RH}|=192$.
For each sample and completed scale, a single Bernoulli draw with
probability $p_{\mathrm{SC}}$ determines whether the entire three-part
ground-truth scale is replaced by these predicted codes. Thus, the SC
decision is shared across all temporal positions and body/hand parts.

GCC is applied after SC. Each valid context token is independently
selected with probability $p_{\mathrm{GCC}}$ and replaced by a uniformly
sampled code from the corresponding part vocabulary. Padding positions
are never corrupted.

Both probabilities are scheduled by optimizer update $s$:
\begin{equation}
\rho(s;p_{\max},w,r)
=
\begin{cases}
0, & s<w,\\
p_{\max}\min\left(1,\frac{s-w}{r}\right), & s\ge w.
\end{cases}
\end{equation}
We use
\begin{equation}
p_{\mathrm{SC}}(s)=\rho(s;0.15,2000,6000),
\qquad
p_{\mathrm{GCC}}(s)=\rho(s;0.12,1000,5000).
\end{equation}
The same probabilities are used at all scales. Neither SC nor GCC is
used during inference, where deterministic argmax decoding is used for
benchmark evaluation.

% ============================================================
% A.3 Training Objectives
% ============================================================
\subsection{Training Objectives}
\label{app:training_objectives}

For scale $m$ and part $p$, token cross-entropy is computed over valid
(non-padding) positions:
\begin{equation}
\mathcal L_{\mathrm{CE}}^{(m,p)}
=
-\frac{1}{N_{m,p}}
\sum_{(i,t)\in\Omega_{m,p}}
\log
p_\theta
\left(
z_{i,t}^{(m,p)}
\mid x_i,Z_i^{(<m)},Z_{i,<t}^{(m)}
\right).
\end{equation}
Token CE uses no label smoothing. We use four temporal divisors
$(8,4,2,1)$ with raw scale weights
\[
(w_1,w_2,w_3,w_4)=(1.25,1.10,1.00,1.25),
\]
and body/left-hand/right-hand coefficients
$(1.0,0.625,0.625)$. The complete training objective is
\begin{equation}
\label{eq:app_training_objective}
\begin{aligned}
\mathcal L
={}&
\frac{1}{4.60}
\sum_{m=1}^{4} w_m
\Big[
\mathcal L_{\mathrm{CE}}^{(m,B)}
+0.625\,\mathcal L_{\mathrm{CE}}^{(m,LH)}
+0.625\,\mathcal L_{\mathrm{CE}}^{(m,RH)}
\Big] \\
&+
0.20\,\mathcal L_{\mathrm{len}}
+
0.03\,\mathcal L_{\mathrm{con}},
\end{aligned}
\end{equation}
where $4.60=\sum_m w_m$ normalizes the scale-weighted token loss.

\paragraph{Length prediction.}
Let $L_i\in\{1,\ldots,100\}$ denote the ground-truth finest-scale token
length. The model predicts a 100-way distribution
$p_{i,k}^{\mathrm{len}}$, with target
\begin{equation}
q_{i,k}
=
(1-\epsilon_{\mathrm{len}})
\mathbf{1}[k=L_i-1]
+
\frac{\epsilon_{\mathrm{len}}}{100},
\qquad
\epsilon_{\mathrm{len}}=0.02.
\end{equation}
The length loss is
\begin{equation}
\mathcal L_{\mathrm{len}}
=
-\frac{1}{B}
\sum_{i=1}^{B}
\sum_{k=0}^{99}
q_{i,k}\log p_{i,k}^{\mathrm{len}}.
\end{equation}
At inference, the finest-scale token length is predicted by the
maximum-probability length class.

\paragraph{Text--motion contrastive loss.}
We obtain $\ell_2$-normalized text and motion embeddings
$t_i$ and $m_i$, respectively. The text representation is mean-pooled
from the language encoder, while the motion representation is constructed
from the ground-truth finest-scale tokens. Body, left-hand, and
right-hand token embeddings are combined with weights
$(0.2,0.4,0.4)$ and temporally mean-pooled. Defining
\begin{equation}
s_{ij}
=
\frac{t_i^\top m_j}{\tau},
\qquad
\tau=0.07,
\end{equation}
we use the symmetric InfoNCE objective
\begin{equation}
\mathcal L_{\mathrm{con}}
=
-\frac{1}{2B}
\sum_{i=1}^{B}
\left[
\log\frac{\exp(s_{ii})}{\sum_j\exp(s_{ij})}
+
\log\frac{\exp(s_{ii})}{\sum_j\exp(s_{ji})}
\right].
\end{equation}
Other examples in the local minibatch serve as negatives.

\paragraph{Optimization details.}
Each part tokenizer uses a temporal convolutional encoder--decoder with
width and latent dimension 512, two $\times2$ temporal
downsampling/upsampling stages, and residual-stack depth 3, yielding a
$4\times$ temporal compression. Each tokenizer is optimized using masked Smooth-$L_1$ reconstruction
loss plus an FSQ commitment loss with weight $0.02$. The
body levels $\{4,4,6\}$ yield 96 indices, while the hand levels
$\{4,6,8\}$ yield 192 indices per hand. The tokenizers are trained for
up to 500 epochs using AdamW with learning rate $2\times10^{-4}$,
gradient clipping at $1.0$, and bfloat16 precision. The generator is
optimized for at most 100,000 updates using learning rates of
$2\times10^{-5}$ for mT5 and $10^{-4}$ for newly introduced modules,
with an effective batch size of 192 on eight NVIDIA A100 GPUs.
Checkpoints are selected by generation validation with early stopping
after ten evaluations without improvement.

% ============================================================
% A.4 Robotic Retargeting
% ============================================================
\subsection{Robotic Retargeting}
\label{app:retargeting}

Following task-space human-to-robot transfer~\citep{signbot,platform},
the recovered SMPL-X upper-body targets are expressed in a
torso-centered coordinate frame and scaled according to the target robot
morphology. Let $\mathbf{y}_t$ denote the elbow, wrist, and orientation
targets at frame $t$. The robot configuration is obtained by solving
\begin{equation}
\label{eq:robot_retargeting}
q_t^\star
=
\arg\min_{q\in\mathcal{Q}}
E_{\mathrm{task}}(q;\mathbf{y}_t)
+
\lambda_{\mathrm{post}}E_{\mathrm{post}}(q)
+
\lambda_{\mathrm{temp}}
\lVert q-q_{t-1}^\star\rVert_2^2,
\end{equation}
where $\mathcal{Q}$ enforces joint limits,
$E_{\mathrm{task}}$ matches the human task-space targets, and
$E_{\mathrm{post}}$ resolves kinematic ambiguity by encouraging
appropriate elbow configurations and bimanual relations. The
preceding-frame solution is used for initialization to promote temporal
continuity.

Hand articulation is retargeted independently by matching
palm-normalized MANO finger geometry to the robot hand under joint-limit
and temporal-continuity constraints. The resulting body and hand
references are temporally regularized and tracked by a platform-specific
low-level controller. Motion channels unsupported by the target
embodiment are omitted during retargeting.

\subsection{Physical Embodiment Evaluation}
\label{app:embodiment_eval}

We quantitatively evaluate the embodiment stage on all 30 available
retargeted sequences, comprising 15 ground-truth (GT) motions and the
corresponding 15 motions generated by \ourmethod, for a total of 11,890
scored retargeting frames.

\paragraph{Metrics.}
Retargeting fidelity is evaluated in the same robot-centered task space
used by the retargeting objective, after applying the corresponding
torso-frame transformation and morphology scaling. For task-space joint
$j$ at frame $t$, we compute
\begin{equation}
    e^{\mathrm{pos}}_{j,t}
    =
    \left\|
        p_j(q_t^\star)-y_{j,t}
    \right\|_2,
\end{equation}
where $y_{j,t}$ is the morphology-mapped target and
$p_j(q_t^\star)$ is the position obtained by forward kinematics from the
retargeted robot configuration. We report elbow and wrist position error,
together with the geodesic orientation error of the wrist.

We further evaluate controller-level tracking using closed-loop MuJoCo
simulation with the same torque-PD controller used by the execution
pipeline. Tracking error is measured between the commanded retargeted
reference and the simulated configuration using task-space errors and
joint-angle MAE. Since encoder-level logs are unavailable for the physical
demonstrations, these values quantify simulated closed-loop tracking rather
than measured hardware tracking.

\begin{table}[t]
    \centering
    \caption{
        \textbf{Quantitative physical-embodiment evaluation.}
        Retargeting measures morphology-transfer error, while execution
        measures closed-loop tracking in physics simulation.
        Results are mean $\pm$ standard deviation across sequences.
    }
    \label{tab:embodiment}
    \setlength{\tabcolsep}{3.8pt}
    \renewcommand{\arraystretch}{1.10}
    \small
    \begin{tabular}{llcccc}
        \toprule
        \textbf{Stage}
        & \textbf{Motion}
        & \textbf{Elbow (cm)}
        & \textbf{Wrist (cm)}
        & \textbf{Wrist ori. ($^\circ$)}
        & \textbf{Joint MAE ($^\circ$)} \\
        \midrule

        \multirow{2}{*}{Retargeting}
        & GT
        & $2.0\pm0.4$
        & $4.6\pm1.4$
        & $22.9\pm10.9$
        & -- \\

        & Ours
        & $2.0\pm0.5$
        & $5.6\pm1.9$
        & $31.3\pm17.7$
        & -- \\

        \midrule

        \multirow{2}{*}{Execution (sim.)}
        & GT
        & $1.6\pm0.5$
        & $3.6\pm0.9$
        & $10.3\pm3.2$
        & $2.1\pm0.7$ \\

        & Ours
        & $0.9\pm0.3$
        & $2.1\pm0.6$
        & $5.9\pm1.6$
        & $1.3\pm0.4$ \\

        \bottomrule
    \end{tabular}
\end{table}

\paragraph{Kinematic feasibility.}
The commanded retargeted configurations remain within the enforced joint
ranges. Because the retargeter explicitly clips configurations to these
ranges, the zero commanded-violation rate is expected; more informatively,
$13.8\%$ of arm-joint entries lie on a joint bound. During closed-loop
simulation, $1.67\%$ of executed joint entries transiently exceed the
configured range, with a maximum excess of $9.5^\circ$, primarily at
wrist-roll and finger joints under self-contact.

\paragraph{Discussion.}
The generated motions exhibit elbow retargeting error identical to GT
motion and only a modest increase in wrist position error
($5.6$\,cm versus $4.6$\,cm), indicating that the generated trajectories
remain compatible with the target robot morphology. Wrist orientation is
the dominant residual, suggesting that orientation-aware retargeting is a
remaining source of embodiment error. Closed-loop simulation further shows
that the resulting references can be tracked with centimeter-level
end-effector error and low joint-angle error.

The archived controller rollouts use a fixed 20-fps source rate, whereas
the original sequences have heterogeneous native frame rates. This does
not affect the frame-indexed retargeting or commanded joint-limit metrics,
but may make the simulated tracking errors optimistic. We therefore use
the simulation results as evidence of controller-level feasibility rather
than measured hardware fidelity.

% ============================================================
% A.5 Evaluation Details
% ============================================================
\subsection{Evaluation Details}
\label{app:evaluation_details}

\paragraph{Geometric metrics.}
Let $P_i$ and $G_j$ denote the predicted and reference 3D joint
positions at frames $i$ and $j$, respectively. For
$p\in\{B,LH,RH\}$, let $K_B=12$ and $K_{LH}=K_{RH}=21$. For DTW-JPE,
body predictions are translation-aligned at the SMPL-X root, while each
hand is aligned at its wrist. Denoting the aligned prediction by
$\widetilde P^{p}_{i,k;j}$, the pairwise cost is
\begin{equation}
c_{\mathrm{JPE}}^{p}(i,j)
=
\frac{1}{K_p}
\sum_{k=1}^{K_p}
\left\|
\widetilde P^{p}_{i,k;j}
-
G^{p}_{j,k}
\right\|_2.
\end{equation}

For DTW-PA-JPE, we instead apply a per-frame similarity Procrustes
transform
\begin{equation}
\mathcal A_{ij}(z)=s zR^\top+t,
\end{equation}
which includes translation, rotation, and uniform scale, and define
\begin{equation}
c_{\mathrm{PA}}^{p}(i,j)
=
\frac{1}{K_p}
\sum_{k=1}^{K_p}
\left\|
\mathcal A_{ij}(P^{p}_{i,k})
-
G^{p}_{j,k}
\right\|_2.
\end{equation}
For the body, the similarity transform is fitted using the SMPL-X
joints and evaluated on the 12 upper-body joints; for each hand, it is
fitted and evaluated on the corresponding 21 points.

For either pairwise cost $c$, standard monotonic DTW determines the
minimum-cost alignment path $\pi^\star$. The reported score is
\begin{equation}
E_{\mathrm{DTW}}^{p}
=
\sum_{(i,j)\in\pi^\star} c^{p}(i,j),
\qquad
E_{\mathrm{hand}}
=
\frac{1}{2}
\left(
E_{\mathrm{left}}+E_{\mathrm{right}}
\right).
\end{equation}
All geometric distances are measured in meters. We report the
mean per-sequence cumulative DTW cost over the test set.

\paragraph{Motion dynamics.}
Dynamics are computed on the native-length z-normalized 133D motion
representation. For feature dimension $d$,
\begin{equation}
\widetilde{x}_{i,t,d}
=
\frac{x_{i,t,d}-\mu_d}
{\max(\sigma_d,10^{-8})}.
\end{equation}
We apply a centered five-frame moving average to obtain
$\bar{\mathbf{x}}_{i,t}$ and define the transition magnitude
\begin{equation}
r_{i,t}
=
\sqrt{
\frac{1}{D}
\left\|
\bar{\mathbf{x}}_{i,t+1}
-
\bar{\mathbf{x}}_{i,t}
\right\|_2^2
},
\qquad D=133.
\end{equation}
The motion threshold is calibrated from ground-truth sequences as
\begin{equation}
\tau
=
Q_{0.20}
\left(
\left\{
r^{\mathrm{GT}}_{i,t}
\right\}_{i,t}
\right).
\end{equation}

Dynamic Degree is the fraction of transitions exceeding this threshold,
macro-averaged across sequences:
\begin{equation}
\mathrm{DynamicDegree}
=
\frac{1}{N}
\sum_{i=1}^{N}
\frac{1}{T_i-1}
\sum_{t=1}^{T_i-1}
\mathbf{1}[r_{i,t}>\tau].
\end{equation}
Because $\tau$ is the 20th percentile of GT transition magnitudes, the
reference value is approximately $0.8$.

For sequence $i$, define its mean motion magnitude as
\begin{equation}
M_i
=
\frac{1}{T_i-1}
\sum_{t=1}^{T_i-1} r_{i,t}.
\end{equation}
Magnitude Ratio is the ratio between the dataset-level mean predicted
and GT magnitudes:
\begin{equation}
\mathrm{MagnitudeRatio}
=
\frac{
\frac{1}{N}\sum_i M_i^{\mathrm{pred}}
}{
\frac{1}{N}\sum_i M_i^{\mathrm{GT}}
}.
\end{equation}
Its reference value is therefore $1$.

For Freeze-tail, let
\begin{equation}
K_i^{\mathrm{tail}}
=
\max
\left\{
k:
r_{i,T_i-j}\leq\tau,\;
j=1,\ldots,k
\right\}
\end{equation}
denote the number of consecutive low-motion transitions at the end of
sequence $i$. We define
\begin{equation}
\mathrm{FreezeTail}
=
\frac{1}{N}
\sum_{i=1}^{N}
\mathbf{1}
\left[
\frac{K_i^{\mathrm{tail}}}{T_i-1}
\geq 0.25
\right].
\end{equation}
Thus, Freeze-tail is the fraction of sequences whose terminal static run
occupies at least $25\%$ of all transitions. Dynamic Degree, Magnitude
Ratio, and Freeze-tail are dimensionless.

\paragraph{Back-translation evaluator.}
For semantic evaluation, we train a separate Hybrid FSQ-T5
back-translator on the official How2Sign splits, comprising 28,691
training, 1,578 validation, and 2,167 test sequences. The evaluator
combines frozen body and bilateral-hand FSQ codes with continuous
four-frame motion statistics through gated fusion and conditions a
Flan-T5-base sequence-to-sequence model. The continuous features capture
the mean, endpoint displacement, and standard deviation within each
four-frame chunk. Auxiliary content-word CTC, semantic-distillation, and
masked-FSQ objectives are used during training.

The checkpoint is selected by validation corpus BLEU-4 and subsequently
frozen for all comparisons. On the held-out How2Sign test split,
back-translation from ground-truth motion achieves BLEU-1, -2, -3, and
-4 scores of $22.45$, $8.59$, $3.95$, and $1.96$, respectively. For
back-translation only, generated motions are linearly resampled to the
corresponding reference length before being passed to the evaluator;
geometry and motion-dynamics metrics are computed at their native
generated lengths. Since the back-translator itself is imperfect, these
scores are used as a comparative measure of semantic consistency rather
than an absolute measure of translation accuracy.

% ============================================================
% A.6 Additional Ablations
% ============================================================
\subsection{Additional Ablations}
\label{app:additional_ablations}

\paragraph{Language-model backbone.}
Table~\ref{tab:lm_backbone_ablation} compares language-model backbones
while keeping the motion representation, generator, and training
protocol unchanged. mT5-Large achieves the lowest errors across all four
geometric metrics, with particularly pronounced improvements in hand
motion. Compared with mT5-Base, it reduces DTW-PA-JPE by $5.9\%$ and
$21.2\%$ for the body and hands, respectively, and DTW-JPE by $4.9\%$
and $11.7\%$.

\paragraph{Sign retrieval.}
Inspired by SOKE~\citep{soke}, we augment the text condition with
retrieved word-level sign tokens. As shown in
Table~\ref{tab:retrieval_ablation}, retrieval improves all four
geometric metrics, with a particularly clear gain in hand motion.
Retrieved signs serve only as auxiliary lexical cues; sentence-level
motion remains generated autoregressively rather than assembled from
isolated sign clips.

\begin{table*}[t]
    \centering

    \begin{minipage}[t]{0.48\textwidth}
        \centering
        \caption{Effect of the language-model backbone on motion generation.}
        \label{tab:lm_backbone_ablation}

        \vspace{2pt}
        \setlength{\tabcolsep}{2.8pt}
        \renewcommand{\arraystretch}{1.12}
        \footnotesize

        \begin{tabular}{lcccc}
            \toprule
            \multirow{2}{*}{\textbf{Backbone}}
            & \multicolumn{2}{c}{\textbf{DTW-PA-JPE} $\downarrow$}
            & \multicolumn{2}{c}{\textbf{DTW-JPE} $\downarrow$} \\
            \cmidrule(lr){2-3}
            \cmidrule(lr){4-5}

            & \textbf{Body}
            & \textbf{Hand}
            & \textbf{Body}
            & \textbf{Hand} \\

            \midrule

            mBART
            & 4.38
            & 1.88
            & 4.24
            & 5.95 \\

            mT5-Base
            & 4.21
            & 1.93
            & 4.07
            & 6.05 \\

            \textbf{mT5-Large}
            & \textbf{3.96}
            & \textbf{1.52}
            & \textbf{3.87}
            & \textbf{5.34} \\

            \bottomrule
        \end{tabular}
    \end{minipage}
    \hfill
    \begin{minipage}[t]{0.48\textwidth}
        \centering
        \caption{Effect of sign retrieval on motion generation.}
        \label{tab:retrieval_ablation}

        \vspace{2pt}
        \setlength{\tabcolsep}{2.0pt}
        \renewcommand{\arraystretch}{1.12}
        \footnotesize

        \begin{tabular}{lcccc}
            \toprule
            \multirow{2}{*}{\textbf{Configuration}}
            & \multicolumn{2}{c}{\textbf{DTW-PA-JPE} $\downarrow$}
            & \multicolumn{2}{c}{\textbf{DTW-JPE} $\downarrow$} \\
            \cmidrule(lr){2-3}
            \cmidrule(lr){4-5}

            & \textbf{Body}
            & \textbf{Hand}
            & \textbf{Body}
            & \textbf{Hand} \\

            \midrule

            w/o Retrieval
            & 4.68
            & 1.91
            & 4.58
            & 6.10 \\

            \textbf{w/ Retrieval}
            & \textbf{3.96}
            & \textbf{1.52}
            & \textbf{3.87}
            & \textbf{5.34} \\

            \bottomrule
        \end{tabular}
    \end{minipage}
\end{table*}

\end{document}